\documentclass[]{interact}

\usepackage{epstopdf}
\usepackage{subfigure}

\usepackage{lineno}
\usepackage{natbib}
\usepackage{multirow}
\usepackage{array}
\usepackage{graphicx}
\usepackage[export]{adjustbox}
\bibpunct[, ]{(}{)}{,}{a}{}{,}
\renewcommand\bibfont{\fontsize{10}{12}\selectfont}

\theoremstyle{plain}

\theoremstyle{definition}

\theoremstyle{remark}

\begin{document}


\title{Segment Anything Model Can Not Segment Anything: Assessing AI Foundation Model's Generalizability in Permafrost Mapping}

\author{
\name{Wenwen Li\textsuperscript{1,$^\ast$}\thanks{$^\ast$CONTACT Wenwen Li. Email: wenwen@asu.edu},  Chia-Yu Hsu\textsuperscript{1}, Sizhe Wang\textsuperscript{1, 2}, Yezhou Yang\textsuperscript{2}, Hyunho Lee\textsuperscript{1}, Anna Liljedahl\textsuperscript{3}, Chandi Witharana\textsuperscript{4}, Yili Yang\textsuperscript{3}, Brendan M. Rogers\textsuperscript{3}, Samantha T. Arundel\textsuperscript{5}, Matthew B. Jones\textsuperscript{6}, Kenton McHenry\textsuperscript{7}, and Patricia Solis\textsuperscript{1}}
\affil{\textsuperscript{1}School of Geographical Sciences and Urban Planning, Arizona State University, Tempe, AZ 85287; \textsuperscript{2}School of Computing and Augmented Intelligence, Arizona State University, Tempe, AZ 85287;
\textsuperscript{3}Woodwell Climate Research Center, Falmouth, MA, 02540;
\textsuperscript{4}Department of Natural Resources and the Environment, University of Connecticut, Storrs, CT, 06269;
\textsuperscript{5}U.S. Geological Survey, Center of Excellence for Geospatial Information Science; Rolla, MO, 65401;
\textsuperscript{6}National Center for Ecological Analysis \& Synthesis, University of California, Santa Barbara, CA, 93106;
\textsuperscript{7}National Center for Supercomputing Applications (NCSA), University of Illinois at Urbana-Champaign, IL, 61820
}
}

\maketitle

\begin{abstract}
This paper assesses trending AI foundation models, especially emerging computer vision foundation models and their performance in natural landscape feature segmentation. While the term foundation model has quickly garnered interest from the geospatial domain, its definition remains vague. Hence, this paper will first introduce AI foundation models and their defining characteristics. Built upon the tremendous success achieved by Large Language Models (LLMs) as the foundation models for language tasks, this paper discusses the challenges of building foundation models for geospatial artificial intelligence (GeoAI) vision tasks. To evaluate the performance of large AI vision models, especially Meta's Segment Anything Model (SAM), we implemented different instance segmentation pipelines that minimize the changes to SAM to leverage its power as a foundation model. A series of prompt strategies was developed to test SAM's performance regarding its theoretical upper bound of predictive accuracy, zero-shot performance, and domain adaptability through fine-tuning. The analysis used two permafrost feature datasets, ice-wedge polygons and retrogressive thaw slumps because (1) these landform features are more challenging to segment than manmade features due to their complicated formation mechanisms, diverse forms, and vague boundaries; (2) their presence and changes are important indicators for Arctic warming and climate change. The results show that although promising, SAM still has room for improvement to support AI-augmented terrain mapping. The spatial and domain generalizability of this finding is further validated using a more general dataset EuroCrop for agricultural field mapping. Finally, we discuss future research directions that strengthen SAM's applicability in challenging geospatial domains.

\end{abstract}


\section{Introduction}
Recent advances in large language models (LLM), such as OpenAI’s GPT (Generative Pre-Trained Transformer), have brought a new wave of important changes to AI. LLMs are trained on millions of web documents (as natural language text) to predict the next or missing words in a sentence, which can be accomplished in a self-supervised manner \citep{devlin2019bert}. These models adopt new architectures and cutting-edge learning strategies \citep[e.g., transformer and multi-head attention;][]{vaswani2017attentiona} and are capable of extracting language structures and patterns, enabling them to handle multiple new tasks, such as question answering and code debugging, for which they were not originally trained. As a result, LLMs have gained a dramatic capability in generalization and language understanding, and they can serve as the foundation for a wide variety of downstream tasks. This is why the term ``foundation model'' was coined in 2019.

What is an AI foundation model? According to \citet{bommasani2021opportunities}, foundation models are trained with massive data to derive an unprecedented ability for knowledge representation, which can further be adapted to solve multiple problems across a diverse set of domains. The technology that enables foundation models to train data at scale is self-supervised learning \citep{devlin2019bert}. In traditional machine learning (ML), feature engineering is required, and this manual process limits ML’s ability to handle big data. The deep learning paradigm in AI enabled automated feature extraction and representation during the model training process. In order to achieve good model generalizability, a large human-annotated dataset was needed. With the emergence of self-supervised learning (for example, contrastive learning and self-distillation), AI models can finally go beyond the limitation of labeled training data and conduct at-scale learning to obtain a more powerful representation from rich, unlabeled big data. A second commonly viewed characteristic of foundation models is that they are large models, consisting of high-depth neural networks with millions to hundreds of millions of parameters and smart learning mechanisms (such as, attention and gating structures) to capture long range, multiplicative interactions within the raw data. A third and perhaps the defining characteristic of foundation models is their ability for domain adaptation, which concerns a model’s ability to conduct pre-training on one data distribution and adapt it to solve a different domain problem through very little extra training effort. When domain adaptation can be achieved with no requirement for training on new datasets, it is called zero-shot learning. When a few samples need to be offered to the model to adapt it for the new task, it is called few-shot learning. Because the foundation models’ goal is to substantially alleviate the labeling cost of supervised learning, their domain adaptation ability has become key to evaluating a model’s generalizability and knowledge transferability.

Embracing foundation models for environmental and social science research has garnered significant interests from the geospatial community. As an emerging concept and exciting new development, foundation models offer the prospect of reducing model development time for individual researchers, and gaining the capability needed to analyze the ever-increasing amount of geospatial data. However, in comparison to the research progress in constructing foundation models for natural language processing (NLP), the advances of vision foundation models are catching up in speed. This is because of the different learning goals between language and vision tasks. Many of the NLP tasks can be formulated as text-to-text processing, which involves taking natural language as input and providing natural language response as output \citep{brown2020language}. When the training data reach a certain size, the LLMs (e.g., ChatGPT) are able to gain incredible generalizability and can be adapted to a diverse set of NLP tasks even without fine-tuning. The success of ChatGPT provides evidence for the power of such LLMs. In contrast, not all computer vision models can be designed as generative models. The goal for vision tasks varies depending on the required granularity in image analysis, be it for visual question answering, image reconstruction, image-level classification, object detection, or instance segmentation. Due to this variability, it is difficult to derive a generalized model for vision tasks, and the emerging vision foundation models, therefore, remain task-specific.

In this research, we aim to assess the strengths and weaknesses of emerging vision foundation model, especially the Segment Anything Model, in its capacity to support GeoAI vision tasks for permafrost mapping. This model was chosen as it is the first publicly-released large vision model for image segmentation and one of the few that are open-sourced and allow for model adaptation using geo-domain data. Two challenging permafrost datasets, ice wedge polygons and retrogressive thaw slumps, were chosen for evaluating SAM's performance in zero-shot prediction, knowledge-embedded learning, as well as its prediction accuracy with model integration and fine-tuning on SAM. The results are compared with those from a cutting-edge model, MViTv2 \citep{li2022mvitv2}, based on supervised learning and multi-scale, transformer-based architecture.

The rest of the paper is organized as follows: Section 2 reviews the growing list of vision foundation models, their model architecture, and supporting vision tasks. Section 3 describes the datasets and a series of strategies we developed to evaluate SAM’s instance segmentation performance, as well as the results. Section 4 summarizes our findings and discusses the strengths and limitations of SAM for AI-augmented permafrost mapping. The extension of this workflow to other geospatial problems, such as agricultural field mapping, is also discussed. Section 5 concludes the paper and propose future research directions.

\section{AI foundation models for GeoAI vision tasks: Recent progress}
As \citet{li2022geoaihsu} pointed out, the evolution of GeoAI may be seen in three phases: (I) import – to bring AI technology into geography and apply it to a domain problem; (II) adaptation – to develop problem-specific strategies to improve general AI models to achieve better performance in various geospatial tasks; (III) export – the integration of geospatial principles and knowledge back to AI to develop innovative models to help better solve both geospatial and aspatial problems. Exciting progress has been made in the past few years, and the field is moving quickly beyond Phase I toward Phase II and III. For adopting new technology, such as foundation models, we still need to go through these three phases, and the study of GeoAI foundation models is clearly at the initial, exploration phase (Phase I). 

Excitingly, there has already been preliminary research toward developing foundation models in the field of remote sensing image analysis. For example, \citet{cha2023billionscale} reported a billion-scale foundation model tailored from the Vision Transformer (ViT) based model. It is constructed with 2.4 billion parameters and pre-trained on the MillionAID dataset \citep{long2021creating}. MillionAID is annotated for image scene classification, with each training image assigned with a scene label, such as dry land, oil field, and sports land. Next, the model is further fine-tuned to support two downstream tasks: rotated object detection and semantic segmentation. The results show that the proposed large model performs better overall than other models, such as RetinaNet \citep{lin2017focala} and Masked Auto Encoding \citep[MAE;][]{wang2022advancing}, on the two remote sensing image processing tasks. It also confirms that when a model is trained on a larger, easier-to-retrieve dataset (MillionAID contains 1 million image scenes) and fine-tuned on other smaller datasets, its performance can be improved as the model becomes more ``knowledgeable'' by learning from bigger data. 

\begin{table}
\tbl{Comparison of data and model operation space for foundation models and (pre) foundation models. NLP: Natural language processing. CV: Computer vision. M: million, B: billion, T: trillion.}
{\begin{tabular}{l l >{\raggedright}p{0.35\textwidth} >{\raggedright}p{0.12\textwidth} c l c} 
 \toprule
 Organization & Model & Data size & Model parameter space & Released date & Task & Open source? \\ 
 \midrule
 Google & BERT & BooksCorpus (800M words) and English Wikipedia (2500M words) & 110M (base) 340M (large) & 2018 & NLP & \checkmark \\
 OpenAI & GPT-3 & 570GB text data from various sources & 175B & 2020 & NLP & - \\
 Google & GLaM & 1.6T tokens & 1.2T & 2021 & NLP & - \\
 DeepMind & Gopher & 300B tokens & 280B & 2021 & NLP & - \\
 Google & PaLM & 780B tokens & 540B & 2022 & NLP & - \\
 OpenAI & GPT-3.5 & 300B tokens & 175B & 2022 & NLP & - \\
 Meta & LLaMA & 1.4T tokens & 65B & 2023 & NLP & \checkmark \\
 OpenAI & GPT-4 & N/A & 1T & 2023 & NLP & - \\
 OpenAI & CLIP & 400M image-text pairs & 150M & 2021 & NLP \& CV & \checkmark \\
 OpenAI & DALL·E & 250M text-image pairs & 12B & 2021 & NLP \& CV & - \\
 Microsoft & Florence & 900M image-text pairs & 893M & 2021 & NLP \& CV & - \\
 Meta & SAM & 11M images with 1B masks & 636M & 2023 & CV & \checkmark \\
 \citealp{cha2023billionscale} & ViT G12x4 & 1M images & 2.4B & 2023 & CV & - \\
  IBM-NASA & Prithvi & 30-meter, 6-band satellite imagery over the continental US & 100M & 2023 & CV & \checkmark \\
\bottomrule
\end{tabular}}
\label{tbl_models_and_data}
\end{table}

In the computer vision field, several big technology organizations, such as Meta, OpenAI, and Microsoft, have put efforts into developing vision foundation models (Table \ref{tbl_models_and_data}). In 2021, OpenAI released CLIP \citep[Contrastive Language-Image Pre-training;][]{radford2021learning} as a transferable visual model. The model learned feature representation by matching pairs of text (captions) and images during the pre-training phase, comparing the similarity between the encoded text and image information. This way, if an image is described as ``a picture of a dog,'' the model will learn the representation of ``dog'' from the images through natural language supervision. Based on this information, CLIP can achieve zero-shot prediction on image scene classes by finding the textural category with the highest similarity to the image's content. CLIP can be adapted to support traditional vision tasks, such as image classification and video action recognition. The model's performance was tested on over 30 vision tasks and the results demonstrated CLIP's comparable zero-shot performance on large benchmark datasets, such as ImageNet and ObjectNet. However, CLIP's performance was less satisfying when analyzing images not included in its pre-training datasets. Hence, the limitation in the size and distribution of training data will directly affect performance in downstream tasks. Furthermore, CLIP cannot conduct fine-granularity tasks, such as object detection and instance segmentation.

Microsoft has developed a computer vision foundation model named Florence, which is also trained on a large set of image-text pairs (900 million) collected from the Internet. Florence aims to achieve zero-shot transfer learning by expanding the representation from coarse to fine-granularity, from static images to dynamic scenes, and from RGB images to images with multiple modalities and channels. This way, Florence can be adapted to more vision tasks than CLIP, including object detection. Florence's pre-training model uses two-tower pipelines, including a CLIP language encoder and a Swin Transformer-based image encoder to encode the image and text data. It uses a contrastive loss function that classifies all image and text description pairs that can be mapped to a unique caption as positive samples and the rest as negative samples. Florence also adopts advanced learning strategies, such as dynamic head, to achieve better adaptation capability for downstream tasks. The model was tested on over 30 datasets on multiple image analysis tasks, including image classification and object detection. The results show that Florence achieves better zero-shot transfer learning than other large visual models, including CLIP. On object detection, it also achieves state-of-the-art performance compared to other cutting-edge models. However, note that object detection with Florence is not zero-shot learning, and the model needs to be fine-tuned for optimal performance. Although promising, Florence is not open-sourced, rendering it difficult to access. Moreover, the model was trained on 512 NVIDIA-A100 GPUs for several days, making training and retraining the model an expensive computational process.

In April 2023, Meta AI released the Segment Anything Model (SAM), which, for the first time, has the power to perform zero-shot image segmentation for more challenging vision tasks. Although AI models such as CLIP and Florence enable vision tasks by associating images with their descriptive text to capture image-scene-level semantics, their ability to identify instance-level semantics within the scene remains weak. SAM enables this capability through a powerful image encoder-decoder-based architecture. A large transformer model is used as the image encoder, and at the decoding phase, the model requires a prompt from the user input to generate the object mask (or object boundary). As shown in Figure 1(a), the input prompt can be a point, a box, or some text to indicate targets of interest. SAM's pre-training went through three stages: (1) supervised training on a small set of images with manually annotated instance masks; (2) semi-automated training by incorporating annotated masks that are ambiguous for SAM; and (3) fully automated training on the entire 11 million image datasets. SAM's advantage is its ability to represent features by learning from a vast number of images, and it can also help accelerate the speed of object annotation for different segmentation tasks through user-machine interaction. Soon after its release, several integrated models \citep{jiang2023conect} were developed to support downstream tasks by incorporating SAM's object segmentation ability. For example, SAM is connected with CLIP to predict each segmented mask’s (object’s) category, such as water. Figure 1 presents the SAM+CLIP integrated model architecture. The SAM's outputs (segmented masks) are fed to CLIP's input, and by providing CLIP a text prompt (e.g., water), CLIP will output all the masks that belong to ``water'' through text-image similarity matching. This way, instance segmentation can be achieved.

Based on the above analysis and considering model maturity, capability, and open-sourceness, in this paper, we select the SAM model to assess its performance in segmenting natural features. Different from other works that apply SAM in remote sensing, our instance segmentation pipelines retain SAM's entire architecture instead of using some of its submodules, such as feature extraction backbones, in the development of the instance segmentation pipeline. This design maximizes the adoption of SAM as a foundation model that emphasizes easy reuse and requires minimal additional effort in model development.

In the next section, we will describe the datasets used, our experimental design, and the various prompt strategies developed to comprehensively assess SAM's instance segmentation performance.

\begin{figure}
\centering
\includegraphics[width=\textwidth]{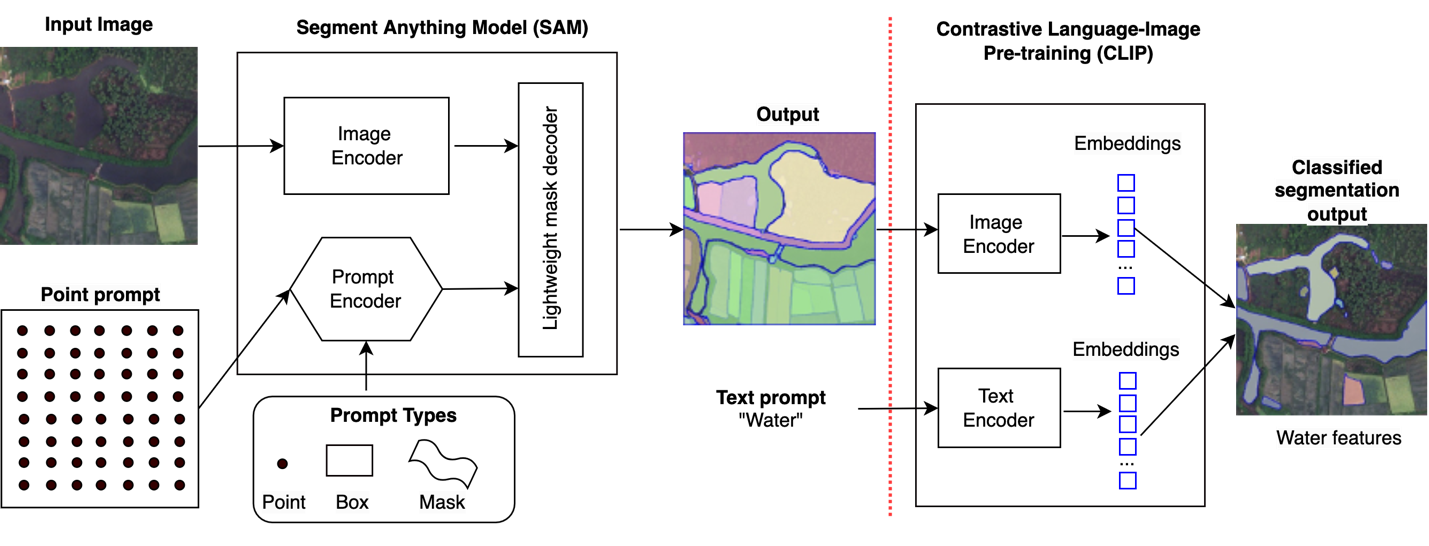}
\caption{Architecture of SAM (left of the dashed line) and CLIP (right of the dashed line) and their combined workflow for instance segmentation.}
\label{fig_sam_clip_arch}
\end{figure}

\section{Data and experiments}\label{data}

\cite{kirillov2023segment} evaluated SAM's performance on multiple vision tasks, including edge detection, object proposal generation, and instance segmentation, all in a zero-shot manner. The results show that, even if SAM was not trained for edge detection, it is capable of generating reasonable edge maps on a benchmark dataset BSDS500 \citep{MartinFTM01}. This provides significant evidence for SAM's task adaptation and transfer learning abilities. For mid-level vision tasks, such as object proposal generation, SAM has exhibited remarkable zero-shot performance in segmenting medium and large objects, outperforming a strong benchmark model ViTDet-H \citep{li2022exploring} based on supervised learning. This experiment was conducted on a challenging vision dataset LVIS (Large Vocabulary Instance Segmentation; \citealp{gupta2019lvis}). Regarding high-level vision tasks, although SAM's zero-shot segmentation accuracy is lower than the supervised model ViTDet-H, the gap is small (8.82\% on the benchmark COCO dataset [\citealp{lin2014microsoft}] and 4.08\% on the LVIS dataset). Furthermore, its mask quality was rated higher than other supervised models based on human studies.

\subsection{Datasets}
While these studies demonstrate SAM's satisfying performance in general computer vision benchmark datasets, its domain adaptation capability for segmenting natural landscape features remains unknown. To address this question, we have selected two natural feature datasets - ice-wedge polygons (IWP) and retrogressive thaw slumps (RTS) - for the assessment. 

The first dataset is the AI-ready Ice Wedge Polygon (IWP) data for instance-level segmentation. IWPs are ambiguous ground surface features found in permafrost-affected landscapes, specifically in regions with ice-rich permafrost. The type of IWP changes when the upper section of an ice wedge melts, indicating the rate of Arctic warming \citep{liljedahl2016panarctic}. Our training dataset covers the most prevalent types of polygonal landscapes found in tundra regions, such as sedge, tussock, and barren. The very high-resolution satellite imagery at 0.5m resolution from commercial Maxar sensors is used as the training image \citep{witharana2020understanding}. 867 image tiles at the sizes between 226 $\times$ 226 to 507 $\times$ 507 were processed. A total of 34,931 IWP were manually annotated within the study area \citep{li2022realtimeb}. 

The second dataset is the AI-ready Retrogressive Thaw Slumps (RTS) data, also curated for instance-level segmentation. Thaw slumps are active permafrost features that develop rapidly when ice-rich permafrost thaws. They are a type of landslide formed when ground ice begins to melt causing the ground to become unstable and the soil to move, especially on steep slopes. This dataset contains 855 image tiles using the 4m Maxar imagery as the base map. A total of 965 RTS were annotated in the study area covering Yamal and Gydan peninsulas in Siberia and six other pan-Arctic sites in Canada and Northeastern Siberia \citep{yang2023mapping}. 

These two features are used to evaluate SAM's image segmentation capabilities for two reasons. First, natural feature datasets are generally more challenging to detect than humanmade features, which are common targets in general AI computer vision tasks and many remote sensing image analysis tasks \citep{li2022geoaihsu}. Their forms are driven by underlying complex geospatial processes, leading to large variations across geographical locations, scales, and landscapes. Because research data (e.g., high-resolution satellite imagery) is often managed in databases and not as readily available as images (e.g., street views) containing humanmade features, their inclusion in existing large pre-trained models can be very limited. For this reason, they also become an ideal dataset to test the domain adaptation capabilities for general-purpose foundation models for GeoAI vision tasks. Second, both IWP and RTS are important permafrost features, the changes of which provide a strong linkage to Arctic warming and climate change \citep{nitze2022towards}. Therefore, AI-augmented permafrost mapping is becoming increasingly important to provide scientific insights into the pace of permafrost thaw and to support global change research, monitoring, and policy \citep{liljedahl2016panarctic, meredith2019polar,natali2022incorporating}.

\subsection{Experimental setup}
We designed a series of experiments to evaluate SAM's potential usage for natural feature segmentation, particularly for important Arctic permafrost features. In an instance segmentation task, a model needs to provide not only the mask indicating the exact boundary of an object, but also a prediction of the object's class. Object localization, object segmentation, and object class prediction are the three factors that affect an instance segmentation model's performance \citep{hsu2021learning}. Because SAM's goal is to segment ``anything,'' its output contains only masks of all objects without their classes. Therefore, SAM, on its own, cannot perform instance segmentation. It must always be combined with other models to create an instance segmentation workflow. 

The first approach is to use SAM to segment objects of any class within an image scene and then connect it with a mask classifier to filter interested objects belonging to a specific class, such as IWP. The SAM and CLIP integrated pipeline shown in Figure \ref{fig_sam_clip_arch} belongs to this category. No training is required in the entire pipeline, except for providing a text prompt, for example, ice wedge polygon, to CLIP. Hence, this process is also known as zero-shot learning (Strategy 1 in Table \ref{tbl_sam_inst_strategies}). 

The second approach is to provide SAM with prior knowledge, for example, the location (a point or a bounding box) of objects of interest as a prompt, and then ask SAM to generate the segmented masks of objects of that class. This is a surrogate of instance segmentation; Strategies 2 to 4 in Table \ref{tbl_sam_inst_strategies} are such examples. Strategy 2 is to feed the ground truth BBOX to SAM and ask SAM to segment the object within the given region. This strategy provides the strongest knowledge among all strategies. Strategy 3 involves feeding SAM with the ground truth point locations for the objects of interest and asking the model to segment the object near the point. Strategy 4 involves training an object detector through supervised learning on the training datasets and using its predicted BBOX to feed the SAM model for instance segmentation. As Strategies 2 - 4 all require the use of training data in the segmentation pipeline, they are not considered zero-shot learning. Since SAM provides code for the model, the pipelines implementing Strategies 2 - 4 can be fine-tuned using domain-specific datasets.

\begin{table}
\tbl{Strategies for enabling instance segmentation capability of SAM.}
{\begin{tabular}{c p{0.4\textwidth} p{0.1\textwidth} l c c}
\toprule
No. & Strategies & \multicolumn{2}{c}{Prior knowledge embedded?} & Zero-shot? & Fine-tunable? \\
\midrule 
1 & Feed SAM with regular points (32$\times$32) & & - & \checkmark & N/A \\
2 & Feed SAM with ground truth bounding box (BBOX) & & \checkmark (strongest) & - & \checkmark \\
3 & Feed SAM with point locations of the features of interest & & \checkmark & - & \checkmark \\
4 & Feed SAM with object detector predicted BBOX & & \checkmark & - & \checkmark \\
\bottomrule
\end{tabular}
}
\label{tbl_sam_inst_strategies}
\end{table}

\subsubsection{Zero-shot instance segmentation (Strategy 1)}\label{sec_samclip}
In this experiment, we investigate the zero-shot capability of SAM to locate and segment natural landscape features. Since SAM generates only masks without any category information, it is incapable of instance segmentation. To address this, we combine SAM with CLIP to perform instance segmentation tasks and use the IWP and RTS datasets to evaluate their performance. CLIP is also a zero-shot model, which evaluates the correlation between a given text prompt and an input image. Taking advantage of this, we can use CLIP to predict the missing category information of each mask produced by SAM. The CLIP model used in this experiment is ViT-B/32, released by OpenAI in 2021 \citep{radford2021learning}. More specifics of the model can be found in Table \ref{tbl_models_and_data}.

\begin{table}
\tbl{Zero-shot performance of the integrated SAM and CLIP model for instance segmentation. mAP: Mean Average Precision. 50 is a threshold to determine the matches between predicted and ground truth masks. When the IOU (Intersection Over Union) of the two is at or above 50\%, the predicted mask is considered a true positive.}
{\begin{tabular}{lllcc} 
 \toprule
 Model & Dataset & Prompt for CLIP & mAP50 (SAM) & mAP50 (SAM + CLIP) \\ 
 \midrule
 \multirow{4}{*}{SAM + CLIP} & \multirow{2}{*}{IWP} & ice-wedge polygon & \multirow{2}{*}{0.073} & 0.117 \\
 & & ice wedge & & 0.104 \\
 \cmidrule{2-5}
 & \multirow{2}{*}{RTS} & thaw slump & \multirow{2}{*}{0.003} & 0.028 \\
 & & retrogressive thaw slump & & 0.007 \\
 \bottomrule
\end{tabular}}
\label{tbl_sam_clip_performance}
\end{table}

We used a regular 32 by 32 grid of evenly distributed points across the image as a prompt for SAM. The output of SAM includes masks generated for all segmented objects. These masks are then utilized to clip the original image, creating a new image that solely contains the specific object, with black markings outside the mask. We then sent this image to CLIP along with the text prompt to identify masks belonging to the given object class. This SAM+CLIP integrated modeling approach (Figure \ref{fig_sam_clip_arch}) implements zero-shot instance segmentation. We evaluated the model on both IWP and RTS datasets using different text prompts for CLIP. As shown in Table \ref{tbl_sam_clip_performance}, the best zero-shot prediction accuracy (measured by mAP50) for the IWP datasets was 0.117 when using ``ice wedge polygon'' as the prompt for CLIP. Using other text prompts, such as ``ice wedge,'' resulted in lower mAP values. 

\begin{figure}
\centering
\makebox[\textwidth][c]
{
\begin{tabular}{ccccc} 
 & Original image & Ground truth & SAM & SAM + CLIP \\ 
 IWP1 & \includegraphics[width=0.25\textwidth,valign=m]{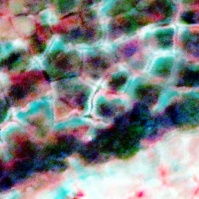} & \includegraphics[width=0.25\textwidth,valign=m]{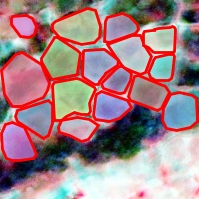} & \includegraphics[width=0.25\textwidth,valign=m]{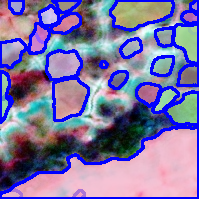} & \includegraphics[width=0.25\textwidth,valign=m]{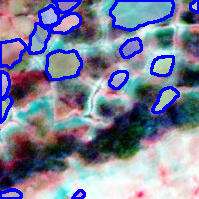} \\
 &&&& \\
 IWP2 & \includegraphics[width=0.25\textwidth,valign=m]{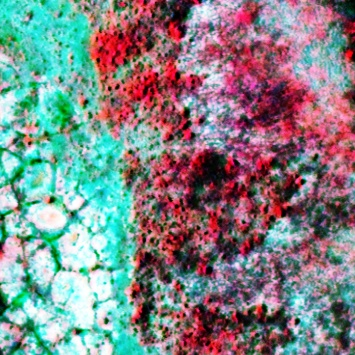} & \includegraphics[width=0.25\textwidth,valign=m]{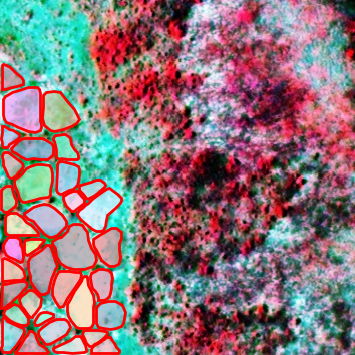} & \includegraphics[width=0.25\textwidth,valign=m]{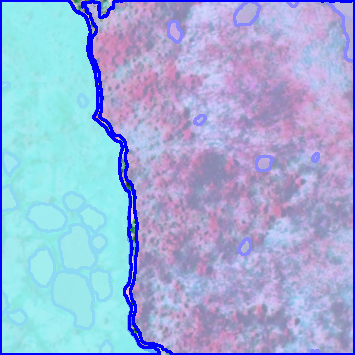} & \includegraphics[width=0.25\textwidth,valign=m]{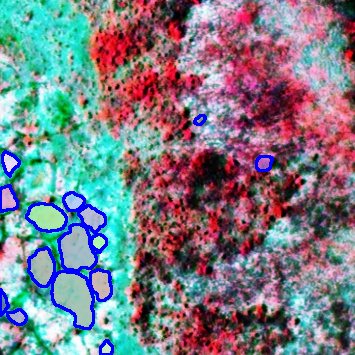} \\
 &&&& \\
 RTS1 & \includegraphics[width=0.25\textwidth,valign=m]{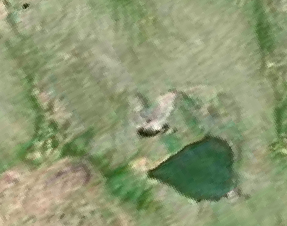} & \includegraphics[width=0.25\textwidth,valign=m]{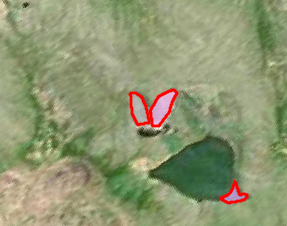} & \includegraphics[width=0.25\textwidth,valign=m]{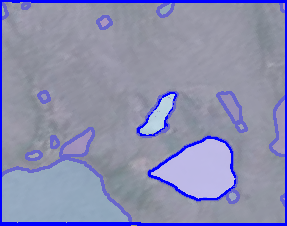} & \includegraphics[width=0.25\textwidth,valign=m]{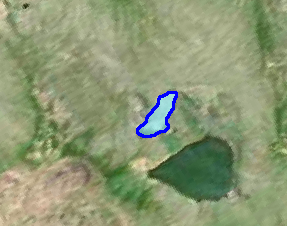} \\
 &&&& \\
 RTS2 & \includegraphics[width=0.25\textwidth,valign=m]{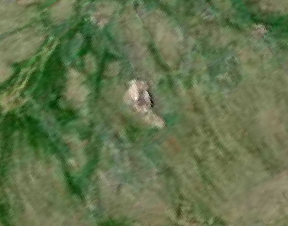} & \includegraphics[width=0.25\textwidth,valign=m]{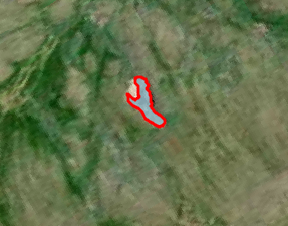} & \includegraphics[width=0.25\textwidth,valign=m]{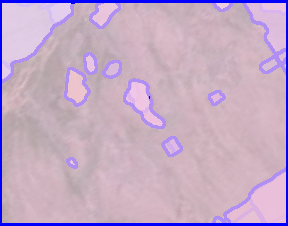} & \includegraphics[width=0.25\textwidth,valign=m]{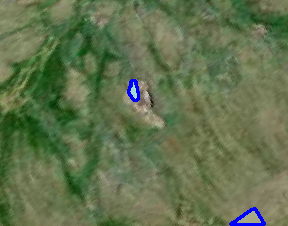} \\
\end{tabular}
}
\caption{Results of zero-shot learning with the integrated SAM+CLIP model. The last column displays the final result, and the second-to-last column presents the intermediate results from SAM, which are used as input for CLIP. Image source: Maxar.com.}
\label{fig_sam_clip_result}
\end{figure}

As indicated by the mAP value, the results are poor. The model's predictions for the RTS dataset were even worse, with a maximum prediction accuracy score of only 0.028. When we examined the results visually (Figure \ref{fig_sam_clip_result}), we discovered that RTS is a more challenging feature to segment than IWP due to its complex and less-defined shape. Since the large SAM/CLIP models do not have enough knowledge of RTS, they tend to have a low success rate. Figure \ref{fig_sam_clip_result} shows two examples (RTS1 and RTS2) of partially segmented RTS features using the SAM+CLIP model, and for many of the testing images, no predictions were made.

Second, since SAM segments all objects within an image scene, it can generate false positives, such as the lower right portion of IWP1 and the two very large regions segmented in IWP2. These false positives can be filtered out by CLIP (last column in Figure \ref{fig_sam_clip_result}). However, CLIP's recall rate cannot be improved beyond SAM's output. Although CLIP can filter out irrelevant masks, it may also filter out true positives. To evaluate the impact of CLIP on SAM's performance, we calculated the mAP50 for SAM alone and the integrated SAM+CLIP model. As shown in Table \ref{tbl_sam_clip_performance}, adding CLIP improved the model's overall performance compared to using SAM alone. This suggests that CLIP did not negatively affect SAM's performance, and the low overall score is due to SAM's limited segmentation capabilities for natural landscape features.

\subsubsection{Knowledge-embedded instance segmentation with SAM (Strategies 2 - 4)}\label{sec_saminst}
Our second set of experiments evaluates SAM's instance segmentation capability using highly accurate and less accurate location prompts. The more accurate location prompts include the ground truth BBOX and the object's center point. Ground truth BBOX provides highly accurate location information about the target objects. Thus the embedded knowledge in these experiments (Strategies 2a and 2b in Table \ref{tbl_sam_performance}) is also the strongest. Masks are not used in these strategies as they provide the answers for SAM. It is important to note that ground truth information (such as ground truth BBOX) should be used only during the training/fine-tuning phase, and the model should not receive any kind of ground truth information during the testing phase, whether for a supervised learning model or a foundation model. Since Strategies 2 and 3 (Table \ref{tbl_sam_performance}) require such ground-truth information to be provided (in the form of a prompt) to SAM during the testing phase, experiments using these strategies indicate only SAM’s theoretical upper-bound segmentation performance and cannot be used in real-world scenarios. In contrast, Strategy 4, which utilizes an object detector to provide BBOX information to SAM for further segmentation, is a practical approach. It involves training the object detector on domain-specific datasets and using it during the testing phase to predict the BBOX of the objects of interest before feeding this information to SAM for object segmentation.

\begin{table}
\tbl{Strategies used to enable SAM's instance segmentation capability through embedded knowledge, and the model's predictive results. Each strategy number corresponds to the definitions provided in Table \ref{tbl_sam_inst_strategies}.}
{\begin{tabular}{p{0.23\textwidth} l c l c c} 
 \toprule
 Model & Learning type & Strategy No. & Prompt & mAP50 (IWP) & mAP50 (RTS) \\ 
 \midrule
 \multirow{3}{*}{SAM without fine-tuning} & \multirow{2}{0.25\textwidth}{Ground truth knowledge plus zero-shot} & 2(a) & Ground truth BBOX & 0.844 & 0.804 \\
 & & 3(a) & Ground truth point & 0.233 & 0.085 \\
 \cmidrule{2-6}
 & supervised learning & 4(a) & Object detector predicted BBOX & 0.521 & 0.290 \\
 \cmidrule{1-6}
 \multirow{3}{*}{SAM with fine-tuning} & \multirow{2}{0.25\textwidth}{Ground truth knowledge plus fine-tuning} & 2(b) & Ground truth BBOX & 0.989 & 0.926 \\
 & & 3(b) & Ground truth point & 0.609 & 0.298 \\
 \cmidrule{2-6}
 & supervised learning & 4(b) & Object detector predicted BBOX & 0.595 & 0.303 \\
 \cmidrule{1-6}
 Benchmark segmentation model \citep[MViTv2;][]{li2022mvitv2} & Supervised learning & 0 & N/A & 0.605 & 0.354 \\

 \bottomrule
\end{tabular}}
\label{tbl_sam_performance}
\end{table}

Because SAM provides the source code for training models, the weights can be further fine-tuned using domain datasets to achieve better results. Our main focus in this experiment was to evaluate the impact of fine-tuning on SAM's performance in segmenting natural features (Strategies 2b, 3b, and 4b in Table \ref{tbl_sam_performance}). SAM comprises three primary components: an image encoder, a prompt encoder, and a mask-decoding transformer, all of which are transformer-based architectures. In our experiment, we froze the model parameters in both the image and prompt encoders and focused our attention on fine-tuning the mask-decoding transformer to reduce computation cost and increase result sensitivity. To optimize the model’s performance, we used Dice Loss, which calculates the overlap between the predicted segmentation masks and the ground truth masks. By comparing the results before and after fine-tuning, we can obtain a more comprehensive understanding of SAM's domain adaptation ability for out-of-distribution datasets.

Table \ref{tbl_sam_performance} shows SAM's strong segmentation performance when provided with ground truth BBOX, achieving mAP50 scores of 0.844 for IWP and 0.804 for RTS (Strategy 2a). However, when given a ground truth point, the accuracy drops significantly to 0.233 for IWP and 0.085 for RTS (Strategy 3a). Interestingly, fine-tuning SAM on domain datasets substantially improves predictive accuracy, with detection accuracy reaching 0.609 for IWP and 0.298 for RTS (Strategy 3b). Despite its poor zero-shot performance on new datasets, SAM demonstrates strong domain adaptation capabilities through fine-tuning  (Table \ref{tbl_sam_clip_performance}). This highlights SAM's generalizability in learning and extracting common feature representations from large image datasets. Proper fine-tuning can significantly enhance its performance on new datasets. However, it is important to note that feeding the model with ground truth information during testing (Strategies 2 and 3) is not practical in real-world scenarios.

Strategy 4, which involves training an object detector and feeding SAM with the predicted BBOX at the testing phase, is a practical solution. Here we selected the Mask region-based convolutional neural network (Mask R-CNN) architecture as our primary object detector because of its dual capability in generating both object BBOX and masks. This allows us to use its predicted BBOX as the input for SAM and also use Mask R-CNN's instance segmentation results as a baseline to evaluate the segmentation quality of SAM. Specifically, we adopted MViTv2 \citep{li2022mvitv2}, a Mask R-CNN type of model that achieves cutting-edge instance segmentation performance in our study. This model uses a multi-scale vision transformer (MViT) as the feature extraction backbone to replace the traditional CNN-based backbone (e.g., ResNet 50) in a Mask R-CNN model. The MViT achieves cutting-edge instance segmentation performance by taking advantage of both the transformer models in capturing long-range data dependencies and the classic CNN models in hierarchical feature learning and strong information flow enabled by residual connections \citep{li2020automateda}. In our experiments, the MViTv2, which has 103M parameters, was adopted as it was tested to yield the best performance on our training datasets. The MViTv2 result also offers a practical upper bound in segmentation accuracy for IWP and RTS. 

Results in Table \ref{tbl_sam_performance} indicate that before fine-tuning, Strategy 4a (feeding SAM with object-detector-predicted BBOX) performs better than feeding SAM with ground truth points (Strategy 3a), but worse than when feeding SAM with ground truth BBOX (Strategy 2a). However, SAM's performance is lower on both datasets when comparing the mAPs of SAM using Strategy 4 to the benchmark segmentation model (MViTv2). After fine-tuning, SAM's predictive accuracy improves from 0.521 to 0.595 (mAP50) for IWP segmentation, which is close to the benchmark model's mAP of 0.605 (Strategy 0). However, for the RTS dataset, SAM's performance does not significantly improve after fine-tuning (0.303 vs. 0.290 for Strategy 4), and a gap remains compared to the benchmark segmentation model's result of 0.354. This gap can be attributed to several factors. First, the RTS dataset is more challenging than the IWP dataset, making the learning of representative features crucial for effective segmentation. The MViT model incorporates innovative strategies to integrate the advantages of CNN, originally designed for image analysis, into transformer models primarily designed for processing sequential data. Consequently, it can capture stronger image feature representations than regular transformer or CNN-based models. Second, our fine-tuning focuses on the decoder part, as fine-tuning the encoder part is very computationally expensive and not effectively feasible.

These experiments show that with and without providing ground-truth BBOX (Strategy 2 vs. Strategy 4), there can be over a 20\% performance gap in the instance segmentation results. Under model fine-tuning and when SAM is fed with ground truth BBOX (Strategy 2b), the model achieves the best segmentation results closest to the ground truth masks. However, in practice, because SAM does not have the ability to perform instance segmentation, it needs to rely on another object detector's predicted results to achieve the goal. Consequently, its performance is limited by the upstream object detector. Therefore, the results for Strategy 4 for both datasets using SAM remain lower than those of the benchmark model MViTv2 (Strategy 0). Additionally, MViTv2's performance can come very close to or be better than the pipeline implementing Strategy 3b when SAM is provided a ground-truth point and fine-tuned using the domain datasets. It is important to mention that Strategies 2 and 3 are not applicable in real-world application scenarios. Hence, the results for Strategy 4 and Strategy 0 have more practical significance. Figure 3 displays segmentation results for the same images shown in Figure 2. These are the results from the fine-tuned SAM and the benchmark segmentation model (MViTv2). 

\begin{figure}
\centering
\makebox[\textwidth][c]
{
\begin{tabular}{l >{\raggedright}p{0.2\textwidth} >{\raggedright}p{0.2\textwidth} >{\raggedright}p{0.2\textwidth} >{\raggedright}p{0.2\textwidth} p{0.2\textwidth}} 
 & Ground truth (GT) & Strategy 2(b) GT BBOX + SAM & Strategy 3(b) GT point + SAM & Strategy 4(b) Ojbect detector predicted BBOX + SAM & Strategy 0 Instance segmentation model (MViTv2) \\ 
 IWP1 & \includegraphics[width=0.2\textwidth,valign=m]{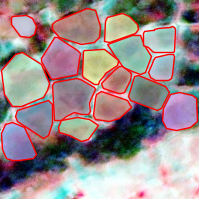} & \includegraphics[width=0.2\textwidth,valign=m]{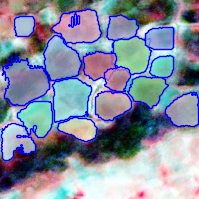} & \includegraphics[width=0.2\textwidth,valign=m]{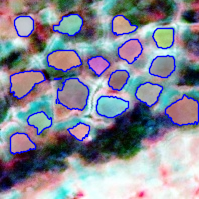} & \includegraphics[width=0.2\textwidth,valign=m]{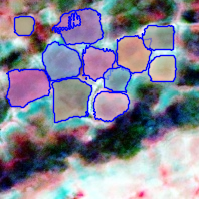} & \includegraphics[width=0.2\textwidth,valign=m]{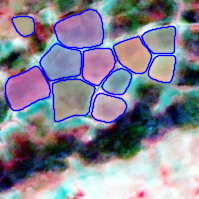} \\
 &&&&& \\
 IWP2 & \includegraphics[width=0.2\textwidth,valign=m]{figures/FID_5048_Polygon_2_gt_bold.png} & \includegraphics[width=0.2\textwidth,valign=m]{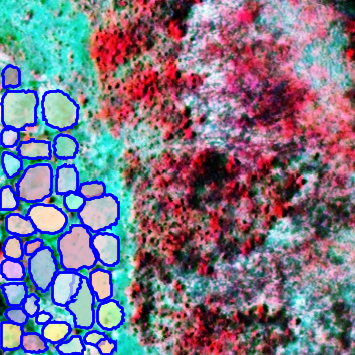} & \includegraphics[width=0.2\textwidth,valign=m]{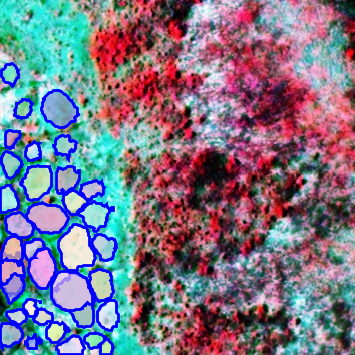} & \includegraphics[width=0.2\textwidth,valign=m]{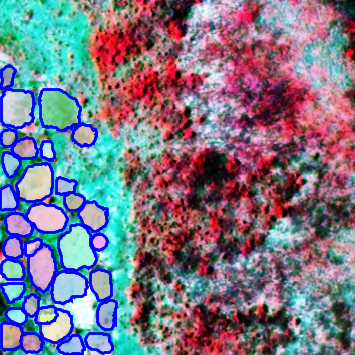} & \includegraphics[width=0.2\textwidth,valign=m]{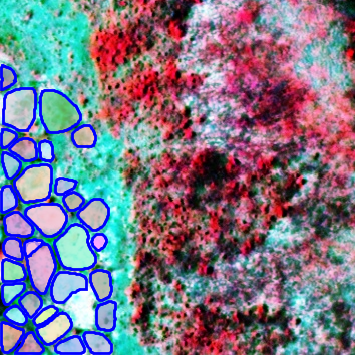} \\
 &&&&& \\
 RTS1 & \includegraphics[width=0.2\textwidth,valign=m]{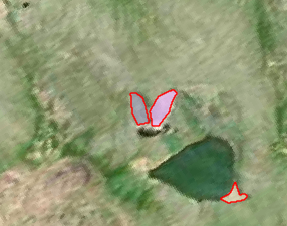} & \includegraphics[width=0.2\textwidth,valign=m]{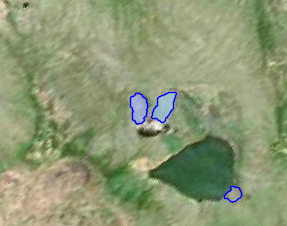} & \includegraphics[width=0.2\textwidth,valign=m]{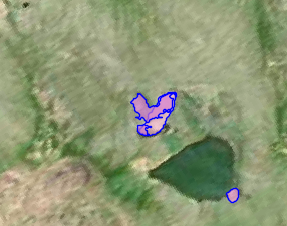} & \includegraphics[width=0.2\textwidth,valign=m]{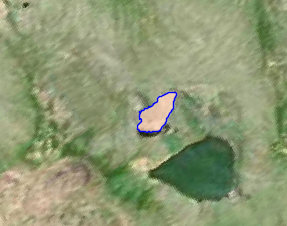} & \includegraphics[width=0.2\textwidth,valign=m]{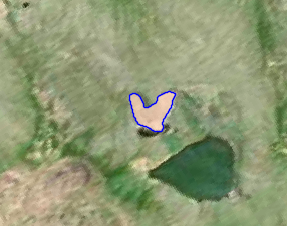} \\
 &&&&& \\
 RTS2 & \includegraphics[width=0.2\textwidth,valign=m]{figures/MAXAR_VALTEST_067_gt_bold.png} & \includegraphics[width=0.2\textwidth,valign=m]{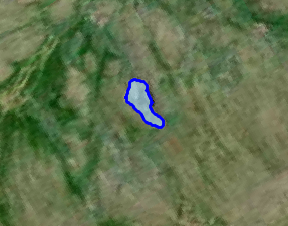} & \includegraphics[width=0.2\textwidth,valign=m]{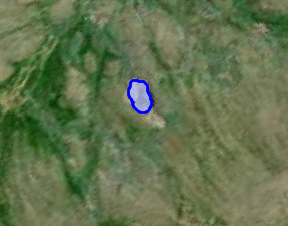} & \includegraphics[width=0.2\textwidth,valign=m]{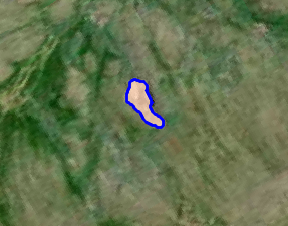} & \includegraphics[width=0.2\textwidth,valign=m]{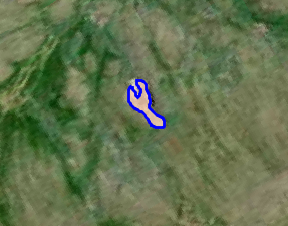} \\
\end{tabular}
}
\caption{Results from knowledge-embedded learning with SAM. The results are those after fine-tuning. The images are the same as those in Figure \ref{fig_sam_clip_result}. Image source: Maxar.com.}
\label{fig_sam_results}
\end{figure}

There are several interesting observations from Figure \ref{fig_sam_results}. First, when SAM is fed with ground truth points (Strategy 3b), the resulting masks tend to be smaller than when SAM is fed a ground truth BBOX (Strategy 2b). This is likely because when the input prompt is smaller, SAM's predictions tend to favor labeling smaller masks with higher confidence. Second, some of SAM's segmentation results exhibit holes (IWP1, Strategy 2b, and Strategy 4b), whereas the MViTv2 results (Strategy 0) do not display any holes. This is likely because SAM is trained on datasets containing holes, whereas MViTv2 is trained exclusively on domain datasets that do not include holes. Third, the results of Strategy 4b (feeding SAM with MViTv2-predicted BBOX) and Strategy 0 (MViTv2) are similar since they both segment masks based on the same BBOX information. Consequently, when MViTv2 fails to make certain predictions (e.g., RTS1), SAM (Strategy 4b) will also be unable to produce the corresponding prediction at the respective location.

\section{Discussions}

\subsection{Summary analysis}
Based on the above analysis, we can see that when directly adopting SAM with no prior knowledge provided (as referenced in the experiment in Section \ref{sec_samclip}), its segmentation capability for out-of-distribution natural features is not good because these features are likely not considered in the model's pre-training processes. Consequently, SAM's zero-shot learning performance for AI-augmented terrain mapping remains poor. Experiments in Section \ref{sec_saminst} show that when providing SAM with strong knowledge - the ground truth BBOX - it can segment the objects inside the target area very well. While ground-truth BBOX cannot be used in a fully automated model prediction scenario, it can be applied in an interactive AI-augmented mapping with human annotators, accelerating the process of data labeling and training data preparation. Some tools have already been developed to integrate SAM into the crowdsourced mapping procedure, such as DS-Annotate \citep{dsannotate}, representing a major step forward in AI and foundation-model-assisted mapping.

When SAM is used as a downstream model to an object detector that trains on the domain dataset and feeds SAM with its predicted BBOX, the model shows much better performance than using it for zero-shot instance segmentation. This is a strategy that can be used in practical applications. However, its performance may be limited by the upstream object detector which feeds it with the predicted BBOX. Comparing SAM’s performance on these natural feature datasets and general-purpose computer vision datasets such as COCO and LVIS (discussed at the beginning of Section \ref{data}), we found that the margins between SAM and a supervised learning model are considerably larger when SAM is adapted for segmenting natural features. Specifically, the difference is nearly 14\% (0.521 for SAM vs. 0.605 for MViTv2 in Table \ref{tbl_sam_performance}) for the IWP dataset and 22\% (0.290 for SAM vs. 0.354 for MViTv2 in Table \ref{tbl_sam_performance}) for the RTS dataset. This suggests a focused area for improving SAM’s domain adaptability in processing geospatial data, especially the challenging permafrost features. 

One positive aspect of SAM is that its performance can be improved after fine-tuning. This is likely due to SAM's strategy of pre-training on a huge number of images, giving it the generalizability to extract common characteristics of diverse objects, even though it might not have learned the intrinsic representation of natural features. Compared to other emerging geospatial vision foundation models, such as IBM's Prithvi which requires 6-band input \citep{Prithvi-100M}, SAM's input is the most commonly seen  RGB bands, hence its has potentially higher adaptability to diverse datasets. 

\subsection{Spatial and domain generalizability test}

To further validate our findings, additional experiments were conducted using SAM for agriculture field mapping. The EuroCrops dataset is employed in this analysis. EuroCrop \citep{schneider2023eurocrops} is the largest harmonized open crop dataset across the European Union; it consists of 944 image scenes captured at the beginning of the growing period in April 2019. The image mosaic is created from two cloud-free Top of Atmosphere (TOA) Sentinel-2 images, each with a spatial resolution of 10 meters. The study area encompasses central Denmark, characterized by dominant agricultural land use and relatively flat terrains. Each image scene measures 128 by 128 in size. 80\% of randomly selected samples are used for training, with the remaining 20\% reserved for testing. 

The same set of experiments was conducted on SAM using this dataset. Table \ref{tbl_sam_clip_agr} illustrates the zero-shot performance by SAM and the combined SAM + CLIP model. It is evident that using SAM alone results in a relatively low model performance. As depicted in Figure 4, SAM may mistakenly segment two crop fields as one large field or fail to identify certain crop fields. Upon incorporating CLIP with SAM and utilizing different prompts for semantic filtering, an increased prediction accuracy is observed, especially when ``Crop land" is used as a prompt. The last column in Figure 4 showcases results using this approach. Notably, CLIP helps to remove some irrelevant urban regions (as seen in image AGR2), contributing to the enhancement of the final results. In contrast, when the prompt ``Agricultural field" is utilized, the model yields slightly lower accuracy than when using SAM alone (Table \ref{tbl_sam_clip_agr}). This suggests that the choice of prompt text is a factor influencing the CLIP model's results.

\begin{table}
\tbl{Zero-shot performance of the integrated SAM and CLIP model for instance segmentation of agricutural fields. The same experimental settings were used as in Table \ref{tbl_sam_clip_performance}.}
{\begin{tabular}{lllcc} 
 \toprule
 Model & Dataset & Prompt for CLIP & mAP50 (SAM) & mAP50 (SAM + CLIP) \\ 
 \midrule
 \multirow{2}{*}{SAM + CLIP} & \multirow{2}{*}{EuroCrop} & Crop land & \multirow{2}{*}{0.118} & 0.161 \\
 & & Agricultural field & & 0.109 \\
 
 \bottomrule
\end{tabular}}
\label{tbl_sam_clip_agr}
\end{table}

\begin{table}
\tbl{Strategies used to enable SAM's instance segmentation capability for agricultural field mapping  through embedded knowledge, and model predictive results. Each strategy number corresponds to the definitions provided in Table \ref{tbl_sam_inst_strategies}.}
{\begin{tabular}{p{0.23\textwidth} l c l c c} 
 \toprule
 Model & Learning type & Strategy No. & Prompt & mAP50 (EuroCrop)  \\ 
 \midrule
 \multirow{3}{*}{SAM without fine-tuning} & \multirow{2}{0.25\textwidth}{Ground truth knowledge plus zero-shot} & 2(a) & Ground truth BBOX & 0.907  \\
 & & 3(a) & Ground truth point & 0.249 \\
 \cmidrule{2-5}
 & supervised learning & 4(a) & Object detector predicted BBOX & 0.685  \\
 \cmidrule{1-5}
 \multirow{3}{*}{SAM with fine-tuning} & \multirow{2}{0.25\textwidth}{Ground truth knowledge plus fine-tuning} & 2(b) & Ground truth BBOX & 0.922 \\
 & & 3(b) & Ground truth point & 0.661  \\
 \cmidrule{2-5}
 & supervised learning & 4(b) & Object detector predicted BBOX & 0.694\\
 \cmidrule{1-5}
 Benchmark segmentation model \citep[MViTv2;][]{li2022mvitv2} & Supervised learning & 0 & N/A & 0.717 \\

 \bottomrule
\end{tabular}}
\label{tbl_sam_performance_agr}
\end{table}

Experiments were also conducted to assess SAM’s performance on additional learning strategies, and the statistical results are shown in Table \ref{tbl_sam_performance_agr}. When SAM is provided with the ground truth BBOX of the agricultural lands, its segmentation accuracy can reach up to 0.907 with zero-shot learning and 0.922 after model fine-tuning. This result is much better than that obtained using a ground-truth point as the input prompt. When used in real-world scenarios and provided with another model's predicted BBOX as input, SAM's prediction accuracy (0.694) can get very close to using a cutting-edge supervised learning model MViTv2 (0.717). Figure \ref{fig_sam_results_agr} illustrates some example results for visual inspection. Overall, the findings on this more general dataset are consistent with what was found in the two permafrost datasets. This further verifies our findings about the strengths and limitations of using SAM in geospatial applications.

\begin{figure}
\centering
\makebox[\textwidth][c]
{
\begin{tabular}{ccccc} 
 & Original image & Ground truth & SAM & SAM + CLIP \\ 
 AGR1 & \includegraphics[width=0.25\textwidth,valign=m]{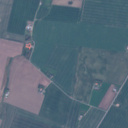} & \includegraphics[width=0.25\textwidth,valign=m]{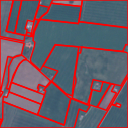} & \includegraphics[width=0.25\textwidth,valign=m]{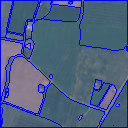} & \includegraphics[width=0.25\textwidth,valign=m]{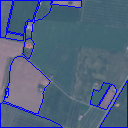} \\
 &&&& \\
 AGR2 & \includegraphics[width=0.25\textwidth,valign=m]{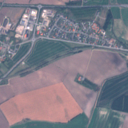} & \includegraphics[width=0.25\textwidth,valign=m]{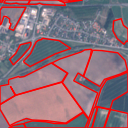} & \includegraphics[width=0.25\textwidth,valign=m]{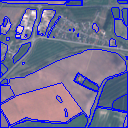} & \includegraphics[width=0.25\textwidth,valign=m]{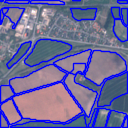} \\
\end{tabular}
}
\caption{Agricultural field mapping results using zero-shot learning with the integrated SAM+CLIP model. The last column displays the final result, and the second-to-last column presents the intermediate results from SAM, which are used as input for CLIP. Image source: Sentinel.}
\label{fig_sam_clip_result_agr}
\end{figure}

\begin{figure}
\centering
\makebox[\textwidth][c]
{
\begin{tabular}{l >{\raggedright}p{0.2\textwidth} >{\raggedright}p{0.2\textwidth} >{\raggedright}p{0.2\textwidth} >{\raggedright}p{0.2\textwidth} p{0.2\textwidth}} 
 & Ground truth (GT) & Strategy 2(b) GT BBOX + SAM & Strategy 3(b) GT point + SAM & Strategy 4(b) Ojbect detector predicted BBOX + SAM & Strategy 0 Instance segmentation model (MViTv2) \\ 
 AGR1 & \includegraphics[width=0.2\textwidth,valign=m]{figures/agr_COCO_val2016_000000100341_gt.png} & \includegraphics[width=0.2\textwidth,valign=m]{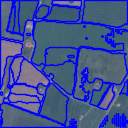} & \includegraphics[width=0.2\textwidth,valign=m]{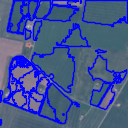} & \includegraphics[width=0.2\textwidth,valign=m]{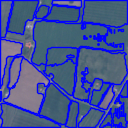} & \includegraphics[width=0.2\textwidth,valign=m]{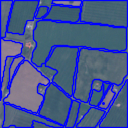} \\
 &&&&& \\
 AGR2 & \includegraphics[width=0.2\textwidth,valign=m]{figures/agr_COCO_val2016_000000100677_gt.png} & \includegraphics[width=0.2\textwidth,valign=m]{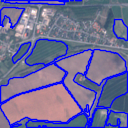} & \includegraphics[width=0.2\textwidth,valign=m]{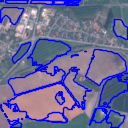} & \includegraphics[width=0.2\textwidth,valign=m]{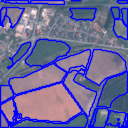} & \includegraphics[width=0.2\textwidth,valign=m]{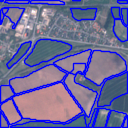} \\

\end{tabular}
}
\caption{Agricultural field mapping results from knowledge-embedded learning with SAM. The results are those after fine-tuning. The images are the same as those in Figure \ref{fig_sam_clip_result_agr}. }
\label{fig_sam_results_agr}
\end{figure}

\section{Conclusions and future research directions}
This paper describes a methodological framework for enabling and assessing vision foundation models for GeoAI vision tasks. Through developing a series of enabling prompt strategies and operational image-analysis pipelines, we have demonstrated multiple pathways to utilize and adopt a new vision foundation model — SAM for GeoAI instance segmentation. The systematic evaluations help us gain an in-depth view of SAM's behavior, as well as its strengths and weaknesses in segmenting challenging permafrost. The set of strategies, including zero-shot learning, model integration, knowledge-embedded learning, and model fine-tuning, can be easily reused and adopted to evaluate SAM's performance across different datasets and support diverse, real-world AI-augmented mapping applications (e.g., agricultural land mapping). 

To emphasize, the instance segmentation pipeline developed in this paper has maximized the use of SAM's entire data processing pipeline instead of using only its submodules, thereby fully utilizing its function as a foundation model and reducing model development costs. Our results show varying performance when providing SAM with different prompts. We have also found a bigger performance gap between SAM and cutting-edge supervised learning-based models in segmenting challenging environmental features (i.e., permafrost), compared to other benchmark computer vision datasets such as COCO and LVIS. This considerable difference emphasizes that SAM's underperformance is more prominent when dealing with natural features (e.g., permafrost) data. However, we have also observed strengths in SAM's model for domain adaptation through the learning of low-level image features shared among different kinds of objects. This strength is evidenced by substantial performance improvement after fine-tuning with permafrost datasets. In the future, more metrics can be further applied to assess SAM's performance in segmenting environmental features, from geometric, spectral, and visual perspectives.

To close the gap between SAM and other cutting-edge supervised learning-based models for instance segmentation of permafrost features (and natural features in general), we suggest several areas of improvement. Data-wise, it is not surprising that SAM's pre-training did not include certain amounts of natural features, as they are often considered as ``background" in an image. Hence, expanding SAM's model with more benchmark natural feature datasets, e.g., \cite{arundel2020geonat} and \cite{hsu2021knowledgedrivena}, will enhance its representation learning ability towards such unique features. Fortunately, SAM's open-source nature enables this expanded capability. Second, these natural features, such as RTS, demonstrate distinct characteristics in different data modalities, such as digital elevation model (DEM) data and Normalized Difference Vegetation Index (NDVI), in addition to optical imagery \citep{wang2021geoaia}. Expanding SAM's base model and allowing it to learn the spectral and statistical properties of natural features will further enhance the model's predictive performance. Third, foundation models, such as SAM, are known for their scalability achieved by training with huge amounts of data. That said, many of the modeling techniques are not new. As AI techniques continue to evolve at an astonishing speed, it is critical to design new model architectures and learning strategies to make the foundation models not only big but also smart at analyzing varying datasets. Hence, it is important for us to continue pushing the edge of fundamental GeoAI research, conducting a comprehensive evaluation of such models regarding their integrity and trustworthiness, and increasing GeoAI (foundation) model transparency and explainability in terms of their design, training, and reasoning process \citep{goodchild2021replicationa}. We also hope that this paper will spark more discussions about adapting SAM and other visual foundation models to support important domain applications within and beyond spatial sciences. 

\section*{Acknowledgement(s)}
This work is supported in part by the National Science Foundation under awards 1853864, 2120943, 2230034, 2230035, and 2033521, as well as Google.org’s Impact Challenge for Climate Innovation Program. Image data access support for this work was provided in part by the Polar Geospatial Center under NSF-OPP awards 1043681 and 1559691.

Any use of trade, firm, or product names is for descriptive purposes only and does not imply endorsement by the U. S. Government.

\bibliographystyle{tfv}
\bibliography{foundationmodel}

\end{document}